\documentclass{article}

\usepackage{microtype}
\usepackage{graphicx}
\usepackage{subfigure}
\usepackage{booktabs} % for professional tables
\usepackage{amsmath}
\usepackage{amssymb}
\usepackage{tablefootnote}
\usepackage{amsthm}
\newtheorem{prop}{Proposition}

\usepackage{bm}
\usepackage{hyperref}

\usepackage[arxiv]{icml2020}

\icmltitlerunning{Graph Pooling with Node Proximity for Hierarchical Representation Learning}

\begin{document}

\twocolumn[
\icmltitle{Graph Pooling with Node Proximity for Hierarchical Representation Learning}

\begin{icmlauthorlist}
\icmlauthor{Xing Gao}{sjtu}
\icmlauthor{Wenrui Dai}{sjtu}
\icmlauthor{Chenglin Li}{sjtu}
\icmlauthor{Hongkai Xiong}{sjtu}
\icmlauthor{Pascal Frossard}{epfl}
\end{icmlauthorlist}

\icmlaffiliation{sjtu}{Shanghai Jiao Tong University, Shanghai, China;}
\icmlaffiliation{epfl}{\'Ecole Polytechnique F\'ed\'erale de Lausanne (EPFL), Lausanne, Switzerland}

\icmlcorrespondingauthor{Xing Gao}{william-g@sjtu.edu.cn}

\icmlkeywords{Machine Learning, ICML}

\vskip 0.3in
]



\printAffiliationsAndNotice{}  % leave blank if no need to mention equal contribution

\begin{abstract}
Graph neural networks have attracted wide attentions to enable representation learning of graph data in recent works. In complement to graph convolution operators, graph pooling is crucial for extracting hierarchical representation of graph data. However, most recent graph pooling methods still fail to efficiently exploit the geometry of graph data. In this paper, we propose a novel graph pooling strategy that leverages node proximity to improve the hierarchical representation learning of graph data with their multi-hop topology. Node proximity is obtained by harmonizing the kernel representation of topology information and node features. Implicit structure-aware kernel representation of topology information allows efficient graph pooling without explicit eigendecomposition of the graph Laplacian. Similarities of node signals are adaptively evaluated with the combination of the affine transformation and kernel trick using the Gaussian RBF function. Experimental results demonstrate that the proposed graph pooling strategy is able to achieve state-of-the-art performance on a collection of public graph classification benchmark datasets.
\end{abstract}

\section{Introduction}
With the advent of data science in various application domains, data is no longer constrained to regular structures, like images and videos, but quite frequently lies on irregular structures represented by graphs (e.g., social networks). Thereby,  a series of pioneer works have been conducted to generalize the state-of-the-art deep learning models used for grid-like data to the hierarchical representation of irregularly structured data. Taking graph data as an example, a collection of  spectral convolution networks \citep{bruna2013spectral,defferrard2016convolutional,khasanova2017graph,kipf2016semi}  and spatial  convolution networks \citep{hamilton2017inductive,wang2018dynamic,velikovi2018graph,zhang2018end,xu2018how,DBLP:conf/aistats/VashishthYBT19} have been developed recently to generalize the convolution operation. On the other hand, even if it is an important module in  multiscale representation learning, the pooling operator for graph neural networks (GNNs) has been mostly overlooked and surely deserves more attention.

Graph pooling  attempts to use a few degrees of freedom (i.e., $|\mathcal{V}|$) to summarize the original graph in terms of both graph topology and graph signal.  One potential solution is to first extract the skeleton of the original graph and then aggregate information of the other graph parts into the skeleton. Intuitively,  the skeleton should strongly coupled with the other nodes in terms of either structure or signal information. Then the assignment of the other graph parts to skeleton nodes should be in accordance with a certain criterion measuring the closeness of different graph parts.   Following these general principles,   several recent works attempt to design differentiable modules  in graph neural networks to extract the skeleton of graphs either explicitly, such as gPool \citep{DBLP:conf/icml/GaoJ19},  SAGPool \citep{DBLP:conf/icml/LeeLK19},  and iPool \citep{gao2019ipool},  or implicitly, e.g., DIFFPOOL \citep{ying2018hierarchical}, and then coarsen the graph.  However, these methods still have some limitations, either in the joint exploitation of the signal and  topology of graph data, or in terms of storage and computational complexity.

Alternatively, spectral graph theory has also provided a large literature on graph analysis, which could potentially lead to graph coarsening schemes. For instance, eigenvalues and eigenvectors associated with a graph are effective in characterizing the topology information of graph data, and several spectral algorithms \citep{von2007tutorial} are proposed to partition graphs. However, there are several limitations of these spectral clustering methods that limit their generalization to the design of graph pooling operators. First,  these methods consider graph topology information but ignore graph signals which contain extra information about the graph data. Furthermore, the  eigendecomposition of the Laplacian matrix associated with a graph is computationally complex and the subsequent k-means clustering algorithm involves iterations. This makes it hard to adopt these operators as  a building block of graph neural networks that should be able to deal with graphs of arbitrary topology simultaneously.

\begin{figure*}[t]
\centering
  \includegraphics[width=6.5in]{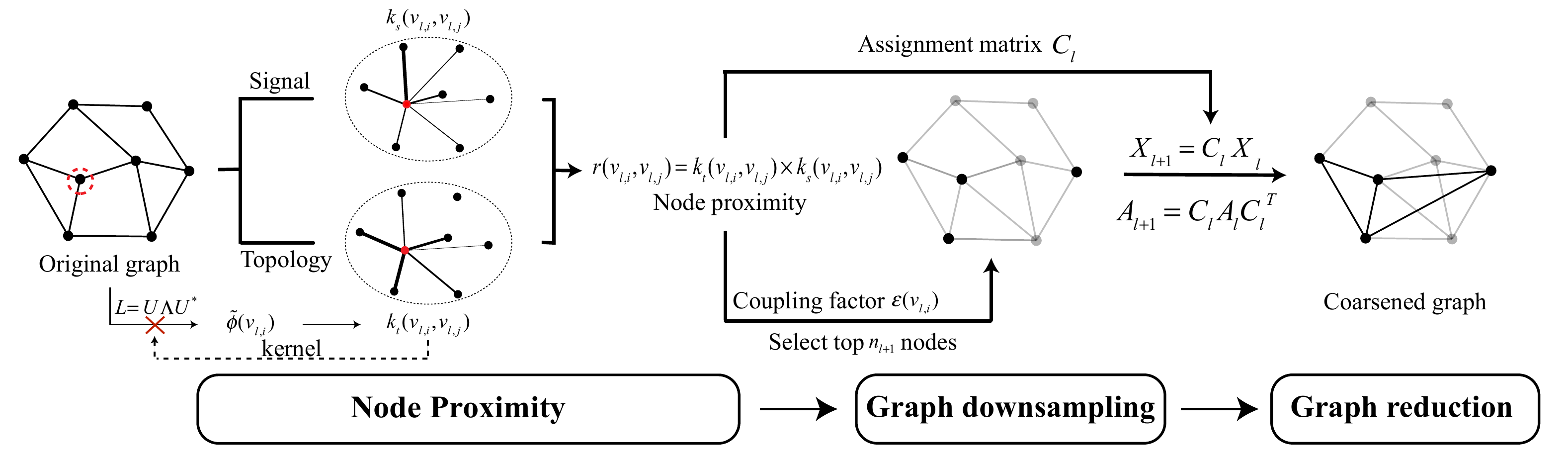}
  \caption{The framework of the proposed pooling operator. For node proximity, a node with the red dashed circle is randomly selected  to illustrate the calculation, where the thickness of the line indicates the proximity between each pair of nodes. On the basis of the node proximity, the coarsened graph is constructed by graph downsampling with seed node selection and graph reduction with  soft-assignment.}\label{fig:1}
\end{figure*}

In this paper, we propose a strategy to generalize the spectral methods to the design of a new graph pooling operator, called ProxPool, by jointly considering the topology and signal information of graph data  without an explicit eigendecomposition. We first introduce a proximity measure to evaluate the closeness of an arbitrary pair of nodes of a graph in terms of both the topology and signal information.  Specifically,  we design a structure-aware kernel on the basis of the eigenvectors of the Laplacian matrix associated with a graph, in order to measure the proximity of nodes that may be not directly connected with an edge. Most importantly, this measure is computed without the need of an explicit eigendecomposition. We further take the signal residing on the vertices of a graph into consideration to characterize the relationship between nodes with an affine transform and a Gaussian RBF kernel. On the basis of the proposed proximity measure that combines signal and topology information, we then propose a novel graph pooling operator consisting of an explicit graph skeleton extraction and a coarsened graph construction, as demonstrated in Fig.~\ref{fig:1}. It is adaptive and flexible as it can deal with pairs of nodes within diverse neighborhood ranges simultaneously. The proposed pooling operator is also interpretable, stackable, and easy to interleave with diverse graph neural networks and can handle graphs of arbitrary structures. It further permits to  achieve state-of-the-art performance on public benchmark graph datasets in terms of graph classification. Our main contributions are as follows:

\begin{itemize}
	\item We present a strategy to measure the closeness between two arbitrary vertices of a graph by jointly considering the signal and topology information of a graph.

	\item To capture the structure proximity between pairs of nodes that are not necessarily connected with a direct edge, we design a structure-aware kernel exploiting spectral properties while avoiding computationally complex eigendecomposition.
	
	\item With our  meaningful node proximity measure, we design an adaptive and stackable graph pooling operator, which permits to achieve state-of-the-art performance on several graph classification benchmark datasets.
\end{itemize}

\section{Related Work}
Several recent methods have been developed for generalizing pooling operators to graphs. For instance, gPool  \citep{DBLP:conf/icml/GaoJ19}  introduces a trainable vector to obtain node footprint and downsamples the graph accordingly. However, gPool ignores the structure information of graphs. SAGPool \citep{DBLP:conf/icml/LeeLK19} coarsens graphs with self-attention. It exploits one-hop structure in graphs by computing attention scores of nodes with a graph convolution operation, which however leads to similar attention scores of nodes within a neighborhood. Thereby, the selected nodes usually concentrate in several specific neighborhoods. DIFFPOOL \citep{ying2018hierarchical} generalizes the Galerkin operator \citep{trottenberg2000multigrid} in algebraic multigrid with  a learnable projection matrix  and obtains the coarsened graph through projections.  However,   the number of parameters of DIFFPOOL depends on the number of vertices, which will impede its applications to large graphs. 

Alternatively, spectral graph theory provides a generalization of the frequency analysis on grid-like data to graph data \citep{shuman2013emerging}. With the spectrum and associated eigenvectors of a graph well characterizing the topology of the graph, a collection of graph processing methods manage to extract hierarchical representation of graphs. For instance, several spectral clustering algorithms \citep{von2007tutorial,biyikoglu2007laplacian} are proposed for graph clustering. A multilevel recursive spectral bisection method is presented in \citep{barnard1994fast}. EigenPooling \citep{DBLP:conf/kdd/0001WAT19} recently designs a graph pooling operator on the basis of Graph Fourier Transform. It relies on the spectral clustering to partition nodes and downsample graphs, and then assigns node signals with the truncated Graph Fourier coefficients of subgraphs. However, it involves high computational complexity and iterations, such as the spectral decomposition, which prevents its uses as a building block of graph neural networks. 

\section{Preliminaries and Framework}
 We consider undirected graphs, and represent them as $\mathcal{G}=(\mathcal{V},\mathcal{E}, A)$. Specifically, $\mathcal{V}$ and $\mathcal{E}$  represent the set of vertices, and the set of edges, respectively. The adjacency matrix $A$ characterizes the graph topology with non-negeative entries, and a non-zero value $(A)_{ij}$ corresponds to an edge connecting vertices $v_i$ and $v_j$, with value one for unweighted graphs or an actual edge weight for weighted graphs.  The degree of vertices is characterized by a degree matrix $D$,   a diagonal matrix with $(D)_{ii}=\sum_{j}(A)_{ij}$. The symmetric normalized Laplacian associated with a graph is defined as $L=I- D^{-\frac{1}{2}}AD^{-\frac{1}{2}}$, and its eigendecomposion  is $L=U \Lambda U^{*}$ with $U=\lbrack \bm{u_1},\bm{u_2}, \dots, \bm{u_{n}} \rbrack$ and $ \Lambda$ being a diagonal matrix composed of $\lambda_1,\lambda_2,\dots,\lambda_{n}$.  Specifically, $\bm{u_i}$ is an eigenvector associated with eigenvalue $\lambda_i$, with $0=\lambda_1 \leq  \lambda_2  \leq \dots \leq   \lambda_{n}\leq 2$. The eigenvalues form the spectrum of the graph and the eigenvectors $\{\bm{u_i}\}$ construct bases of the so called Graph Fourier Transform \citep{shuman2013emerging}. 
 
 Generally,  capital letters represent matrices, while bold lowercase letters  indicate vectors. Furthermore, we employ a subscript $l$ to indicate variables or parameters belonging to the $l$-th layer of the neural network architecture.  For instance, for the graph $\mathcal{G}_{l}=(\mathcal{V}_{l},\mathcal{E}_{l},A_l )$ with $n_l$ vertices in the $l$-th neural network representation layer,  $\bm{x_{l,i}} = x(v_{l,i}) \in \mathbb{R}^{d_l}$ indicates the signal of dimension $d_l$ residing on node $v_{l,i}$, and  $X_l=[\bm{x_{l,1}},\bm{x_{l,2}}, \cdots, \bm{x_{l,n_l}}]^{T}$ represents the signal of the whole graph $\mathcal{G}_l$. 

In this paper, we rely on graph convolution networks (GCNs), which usually consist of a stack of interlaced graph convolution layers and graph pooling layers,  in order to extract multiscale representations of graph data. 
For the spatial convolution networks,  the graph convolution operator usually adopts a neighborhood message aggregation as:
\begin{equation}\label{e:0}
	X_{l+1}=\sigma(\xi(S_{l},X_{l},W_{l})),
\end{equation}
where $S_{l}$ is a graph shift operator (e.g., $A_{l}$ or $L_{l}$),  $\xi(\cdot)$ indicates an aggregation function,  $\sigma(\cdot)$ denotes a non-linear  activation function, and $W_{l}$ are learnable parameters.  The graph pooling operator takes as input the  adjacency matrix and the node signals of the original graph and generates those of the coarsened graphs. It can be generally formulated as
\begin{gather}
    X_{l+1}=C_{l} X_{l},\quad
    A_{l+1}=C_{l} A_l C^T_l,
\end{gather}
where $C_l\in\mathbb{R}^{n_{l+1}\times n_{l}}$ denotes a coarsening matrix, which is the core of the design of graph pooling operators.

\section{Node  Proximity}\label{sec:prox}
We will first introduce a structure-aware kernel to characterize the connection strength between vertices of a graph in terms of graph topology. A RBF kernel further deals with signals residing on the vertices of the graph and complements the structure kernel with the information of node signals. Both kernels are used together to measure the proximity  between pairs of nodes of a graph. 

\subsection{Structure-aware Kernels}
The topology of a graph describes the relationship between nodes and is completely characterized by the adjacency matrix $A$ associated with a graph. With an adjacency matrix, we can obtain the direct connections between vertices but are unable to directly measure the closeness of two nodes without direct edges. In many situations, however, it is desirable to evaluate the closeness of nodes that are not only direct neighbors but connected within an $s$-hop neighborhood. In this section, we introduce a strategy to measure the closeness of nodes within an $s$-hop neighborhood in terms of the topology of the graph.

We propose to resort to a proxy smooth graph signal to evaluate the node proximity for a graph with arbitrary topology. If a graph signal is smooth,  signals that reside on nearby vertices are similar, and we can then measure the closeness of vertices by computing the similarity of the signals they support. We can construct such smooth graph signals for arbitrary graphs with the help of the eigenvectors of their Laplacian matrices.

Specifically, for a symmetric normalized Laplacian matrix $L_l$ associated with a graph $\mathcal{G}_l$, we have
\begin{equation}\label{e:3}
\lambda_{l,k}=\bm{u}_{l,k}^*L_l\bm{u}_{l,k}=\frac{1}{2}\sum_{i,j=1}^{n} (A_l)_{ij}\left(\frac{u_{l,k,i}}{\sqrt{d_{l,i}}}-\frac{u_{l,k,j}}{\sqrt{d_{l,j}}}\right)^2,
\end{equation}
where $\lambda_{l,k}$ and $\bm{u_{l,k}}$ are respectively an eigenvalue and its corresponding eigenvector of  $L_l$. Let us now consider a signal $\bm{\alpha}_{l,k}=D_l^{-\frac{1}{2}}\bm{u}_{l,k}$ with a real value $\alpha_{l,k,i}=u_{l,k,i}/\sqrt{d_{l,i}}$ to each vertex $v_{l,i}$ of $\mathcal{G}_l$.
Equation~\eqref{e:3} shows that $\bm{\alpha_{l,k}}$ is a smooth signal for $\mathcal{G}_l$, when $\lambda_{l,k}$ is small. Note that $\bm{\alpha}_{l,1}=c\bm{1}$ is a vector of constant $c$ that corresponds to the eigenvalue $\lambda_{l,1}=0$. When only the first $m$ eigenvalues are kept, we can further obtain an $m$-dimensional smooth node signal $\phi(v_{l,i})=[\alpha_{l,1,i},\alpha_{l,2,i}, \cdots, \alpha_{l,m,i}, 0, \dots, 0 ]^T \in \mathbb{R}^{n_l}$ for $v_{l,i}$.
Here, $m$ depends on the spectrum of a graph. However, $m$ would not be adaptively found, as the spectrum varies with the diverse topologies of different graph data. Alternatively, we present a universal strategy to construct smooth graph signals. Instead of explicitly determining $m$, we introduce a monotonically decreasing real function $f: [0,2] \rightarrow [0,1]$ with $f(0)=1$ and $f(\lambda) \rightarrow 0$ at $\lambda \rightarrow 2$. For arbitrary vertex $v_{l,i}$, its smooth node signal $\tilde{\phi}(v_{l,i})$ is obtained by suppressing the amplitudes of eigenvectors corresponding to the large eigenvalues with $f(\lambda)$.
\begin{align}\label{e:f}
\tilde{\phi}(v_{l,i})&=[\tilde{\alpha}_{l,1,i},\tilde{\alpha}_{l,2,i}, \cdots, \tilde{\alpha}_{l,n,i}]^T\nonumber\\
&=[f(\lambda_{l,1})\alpha_{l,1,i}, f(\lambda_{l,2})\alpha_{l,2,i}, \cdots, f(\lambda_{l,n})\alpha_{l,n,i}]^T
\end{align}
In Proposition \ref{p:1}, we prove that the dimension of $\tilde{\phi}(v_{l,i})$ equals to the number of eigenvalues satisfying $f(\lambda_{l,i})>0$.

\begin {prop}\label{p:1} Given a graph $\mathcal{G}_l$ without isolated vertices and a monotonically decreasing real function $f: [0,2] \rightarrow [0,1]$ with $f(0)=1$ and $f(\lambda) \rightarrow 0$ at $\lambda \rightarrow 2$, when $t$ eigenvalues of $\mathcal{G}_l$ satisfy $f(\lambda_{l,i})>0$ for $i=1,\cdots,t$, the smooth node signal $\tilde{\phi}(v_{l,i})$ of $v_{l,i}$ defined in Eq.~\eqref{e:f} 
locates in a $t$-dimensional subspace.
\begin{proof}
$(D_l)_{ii}>0$ for $\mathcal{G}_l$ without isolated vertices, which means that $D_l$ and $U_l=[\bm{u}_{l,1},\bm{u}_{l,2},\dots,\bm{u}_{l,n}]$ are invertible. Thus, $\{\bm{\alpha}_{l,1},\bm{\alpha}_{l,2},\dots,\bm{\alpha}_{l,n}\}$ is linearly independent, as $D_l^{-\frac{1}{2}}U_l=[\bm{\alpha}_{l,1},\bm{\alpha}_{l,2},\dots,\bm{\alpha}_{l,n}]$ is invertible. Since $f(\lambda_{l,1})\geq f(\lambda_{l,2})\geq\cdots\geq f(\lambda_{l,t})>0$, $\{\tilde{\bm{\alpha}}_{l,1},\tilde{\bm{\alpha}}_{l,2},\dots,\tilde{\bm{\alpha}}_{l,t}|\tilde{\bm{\alpha}}_{l,i}= f(\lambda_{l,i})\bm{\alpha}_{l,i}\}$ is also linearly independent. As a result, $\tilde{\phi}(v_{l,i})=[\tilde{\alpha}_{l,1,i},\tilde{\alpha}_{l,2,i},\cdots,\tilde{\alpha}_{l,t,i}, 0,\cdots,0]^T$ is a $t$-dimensional signal for $f(\lambda_{l,k})=0$, $t<k \leq n_l$.
\end{proof}
\end {prop}

Proposition~\ref{p:1} implies that the dimension of $\tilde{\phi}(v_{l,i})$ depends on the function $f(\cdot)$ and the spectrum of graphs. Thus, an adaptive strategy is presented to determine $m$. Similarity measure of the smooth node signals enables the fast implementation of the proximity measure without eigendecomposition. The cosine function is widely used to evaluate similarity of variables and its inner-product formulation enables the kernel-based implementation of the proposed proximity measure. The node proximity can be quantitatively measured with the similarity of the proxy smooth graph signal residing on two vertices.
\begin{equation}
k_t(v_{l,i},v_{l,j})=\cos(\tilde{\phi}(v_{l,i}),\tilde{\phi}(v_{l,j}))=\frac{\tilde{\phi}(v_{l,i})^T\tilde{\phi}(v_{l,j})}{\|\tilde{\phi}(v_{l,i})\|_2\|\tilde{\phi}(v_{l,j})\|_2}.
\end{equation}
With the proxy graph signal $\tilde{\Phi}_l=[\tilde{\phi}(v_{l,1}),\tilde{\phi}(v_{l,2}) \dots, \tilde{\phi}(v_{l,n})]^T$, we can obtain this proximity of all the pairs of nodes:
\begin{equation}\label{e:4}
\begin{split}
	K_{t}&=\tilde{D}_l^{-\frac{1}{2}}\tilde{\Phi}_l\tilde{\Phi}_l^*\tilde{D}_l^{-\frac{1}{2}} \\
	&=\tilde{D}_l^{-\frac{1}{2}}D_l^{-\frac{1}{2}}U_lf(\Lambda_l)f^*(\Lambda_l) U_l^*D_l^{-\frac{1}{2}}\tilde{D}_l^{-\frac{1}{2}}\\
	 &=\tilde{D}_l^{-\frac{1}{2}}D_l^{-\frac{1}{2}}U_lg(\Lambda_l) U_l^*D_l^{-\frac{1}{2}}\tilde{D}_l^{-\frac{1}{2}},
\end{split}
\end{equation}
where $\tilde{D}_l$ is a diagonal matrix with the same diagonal values of $\tilde{\Phi}_l\tilde{\Phi}_l^*$ and $g(\Lambda_l)$ indicates a filter in frequency domain with
\begin{equation}
	g(\lambda)=f(\lambda)f(\lambda)^*.
\end{equation}
We choose the filtering function $g(\cdot)$ as 
\begin{equation}\label{e:8}
	g(\lambda)=(1-\frac{1}{2}\lambda)^s.
\end{equation}
Eq.~\eqref{e:8} shows that $g(0)=1$, $g(2)=0$, and $g(\lambda)\rightarrow 0$ decays rapidly with $\lambda\rightarrow 2$ given large $s$. Equation~\eqref{e:4} can be rewritten using this polynomial formation:
\begin{equation}\label{e:9}
	K_{t}=\tilde{D_l}^{-\frac{1}{2}}D_l^{-\frac{1}{2}}g({L_l})D_l^{-\frac{1}{2}}\tilde{D}_l^{-\frac{1}{2}}.
\end{equation}
Eq.~\eqref{e:9} suggests that $K_{t}$ can be obtained without explicit computation-intensive eigendecomposition of the graph Laplacian. Notably, the proposed proximity measure is a generalization of the cluster kernel defined in \citep{chapelle2003cluster}, as $K_{t}$ is a gram matrix that naturally derives a kernel. In comparison to the adjacent matrix $A$, $K_t$ characterizes the topology of the graph with the $s$-hop connections in addition to a direct edge. The proposed proximity measure is also distinctive from the normalized spectral clustering \citep{shi2000normalized} that adopts the node signal $\phi(v_{l,i})=[\alpha_{l,0,i},\alpha_{l,1,i}, \cdots, \alpha_{l,m,i}, 0, \dots, 0 ]^T$ for $k$-means clustering of vertices. We exploit a kernel method to measure the closeness between nodes to avoid complex eigendecomposition and excessive iterations.

\subsection{Node Signal Proximity}

We now measure the proximity of nodes in terms of graph signals. The radial basis function kernel (RBF kernel) is an effective method to calculate similarity of two variables with an implicit mapping. Rather than directly applying RBF kernel, we first project node signals to a low-dimension subspace with an affine transform, which permits to focus on the specific components of the signals:
\begin{equation}\label{e:13}
	Q_{l}=X_lW_{Q,l},
\end{equation}
where $W_{Q,l} \in \mathbb{R}^{d_l \times d^{'}_l }$ is learnable with $d^{'}_l < d_l $ to reduce computation and $Q_{l}=[\bm{q_{l,1}},\bm{q_{l,2}},\dots,\bm{q_{l,n_l}}]^T$.
Then node proximity is computed with a Gaussian RBF kernel:
\begin{equation}\label{e:11}
\begin{split}
	k_{s}(v_{l,i},v_{l,j})&=e^{-\tau \parallel \bm{q_{l,i}}- \bm{ q_{l,j}} \parallel^2_2}\\
	&=e^{ -\tau (\bm{x_{l,i}}- \bm{x_{l,j}})^TW_{Q,l}W_{Q,l}^T(\bm{x_{l,i}}- \bm{x_{l,j}})},
\end{split}
\end{equation}
with $\tau$ as the precision.
We can obtain the corresponding gram matrix $K_s$ that captures all the proximity values between pairs of nodes with
\begin{equation}
	(K_{s})_{ij}=k_{s}(v_{l,i},v_{l,j}).
\end{equation}

In this way, with the affine transform and the implicit mapping of the kernel trick, we are able to adaptively characterize the relationship of nodes in terms of node signals.

\subsection{Node Proximity}
We finally present a strategy to measure node proximity by jointly considering the topology and signal information. In general, two nodes have high proximity when they have a tight interconnections in topology and the node signals they support are closely related. With structure-aware kernels and RBF kernels that respectively measure the proximity of nodes in terms of the topology and signal information, we implement this ``AND Gate" and design the proximity measure by reconciling them with a multiplication: \begin{gather}
        r(v_{l,i},v_{l,j})=k_t(v_{l,i},v_{l,j}) \times k_s(v_{l,i},v_{l,j}); \label{e:12}
\end{gather}
for the whole graph $\mathcal{G}_l$, we have
\begin{equation}
R_l=K_{t}  \odot K_{s}, \label{e:17}
\end{equation}
where $\odot$ indicates the Hadamard product. The contribution of the topology and signal information can be indirectly but adaptively adjusted with the hyper-parameter $\tau$ in Eq.~(\ref{e:11}). Furthermore, the proximity measure $R_l$ can be adapted to exploit local and global information with the choice of neighborhood $s$ in the structure-aware kernel $K_t$.

\section{ProxPool}\label{sec:pool}
On the basis of node proximity, we design a novel graph pooling operator. We first introduce a strategy to evaluate the coupling strength of a vertex with other nodes and then present a graph downsampling operation with proximity-based seed node selection. Finally, we present a graph reduction strategy with soft-assignment to construct coarsened graphs towards hierarchical graph representation. 

\subsection{Graph Downsampling with Seed Node Selection}
We adopt a similar downsampling method as  \cite{zhang2018end,DBLP:conf/icml/GaoJ19,DBLP:conf/icml/LeeLK19}. To faithfully represent the original graph, the nodes selected for graph downsampling should be ``strongly coupled" with other nodes or sufficiently representative.  

Since our node proximity criterion reconciles the topology and signal information, we can use it to govern the seed node selection doing graph downsampling. Specifically, we define the coupling factor for node $v_{l,i}$ as
\begin{equation} \label{e:18}
	\varepsilon(v_{l,i})= \sum_{v_{l,j} \in \mathcal{V}_l \backslash v_{l,i}} \frac{r(v_{l,i},v_{l,j})}{\sum_{v_{l,k} \in \mathcal{V}_l \backslash v_{l,j}}r(v_{l,k},v_{l,j})},
\end{equation}
where $r(v_{l,i},v_{l,j})$ is the proximity measure defined in Eq.~(\ref{e:12}).
$\varepsilon(v_{l,i})$ measures the volume that $v_{l,i}$ gets from other nodes, and thereby the strength of its coupling. With normalization, we break the symmetry of the proximity of each pair of nodes, and the node with more extensive connections with other nodes different than $v_{l,i}$ (i.e., $\sum_{v_{l,k} \in \mathcal{V}_l \backslash v_{l,i}}r(v_{l,k},v_{l,i}) > \sum_{v_{l,k} \in \mathcal{V}_l \backslash v_{l,j}}r(v_{l,k},v_{l,j})$)  gains a greater coupling value ($\frac{r(v_{l,i},v_{l,j})}{\sum_{v_{l,k} \in \mathcal{V}_l \backslash v_{l,j}}r(v_{l,k},v_{l,j})}>\frac{r(v_{l,j},v_{l,i})}{\sum_{v_{l,k} \in \mathcal{V}_l \backslash v_{l,i}}r(v_{l,k},v_{l,i})}$) , which 
eventually favors the selection of strongly connected seed nodes.

For the whole graph, the coupling factor is reformulated as:
\begin{equation}
	E(\mathcal{G}_l)=(R_l-{\rm diag\_embed}(R_l))D_{R_l}^{-1}\bm{1},
\end{equation}
where ${\rm diag\_embed}$ returns a diagonal matrix with the diagonal elements of the input matrix $R_l$ of Eq.~(\ref{e:12}), $D_{R_l}$ is a diagonal matrix with $(D_{R_l})_{ii}=\sum_{j \neq i} (R_l)_{ij}$, and $\bm{1}$ indicates a vector of constant 1.

On the basis of the coupling factor of each vertex, we can finally select seed nodes that are ``strongly coupled'' with others by  re-ordering vertices  in terms of $E(\mathcal{G}_l)$ and keeping the top $n_{l+1}$ ones accordingly, i.e.,
\begin{gather}
    	\textbf{idx}={\rm rank}(E(\mathcal{G}_l), n_{l+1}), 
\end{gather}
where  $\rm rank$ is a global ranking operator that returns the top  $n_{l+1}$  nodes with the largest score and $n_{l+1}=\rho \times |\mathcal{V}_l|$ is dependent on the pooling ratio $\rho$ and the number of nodes.

\begin{table*}[htp]
\centering
 \caption{Results of graph classification in terms of accuracy and standard variation with 20 random data splits.} \label{t:2}
\begin{tabular*}{1.99\columnwidth}{@{\extracolsep{\fill}}l|ccccc}
	\toprule
Dataset	 &     D\&D   &   PROTEINS & NCI1 & NCI109 & MUTAGENICITY\\
	 \midrule
     DIFFPOOL&79.19 $\pm$ 3.35&74.96 $\pm$ 4.14&74.62 $\pm$ 2.04&74.60 $\pm$ 1.88&78.55 $\pm$ 1.87\\
	 gPool&79.32 $\pm$  4.07&74.78 $\pm$ 4.02&75.64 $\pm$ 2.47&75.54 $\pm$ 2.00&80.33 $\pm$ 1.54\\
	 SAGPool&79.06 $\pm$ 3.96&75.09 $\pm$ 4.82&76.23 $\pm$ 2.26&75.80 $\pm$ 2.35&79.99 $\pm$ 2.09 \\
	 EigenPool&78.89 $\pm$ 3.95&75.09 $\pm$ 3.51&76.57 $\pm$ 2.79&76.16 $\pm$ 1.94&80.10 $\pm$ 2.03\\
	 \midrule
	 ProxPool-NT&79.19 $\pm$ 3.49&75.09 $\pm$ 4.09&77.77 $\pm$ 2.04&76.02 $\pm$ 2.04&80.41 $\pm$ 2.11\\
     ProxPool-NS&\textbf{80.17 $\pm$ 2.27}&75.36 $\pm$ 4.22&\textbf{77.87 $\pm$ 2.54}&76.69 $\pm$ 2.67&80.71 $\pm$ 1.78\\
     ProxPool&79.83 $\pm$ 3.18&\textbf{75.67 $\pm$ 4.61}&77.83 $\pm$ 2.39&\textbf{77.02 $\pm$ 1.97}&\textbf{80.71 $\pm$ 1.72}\\
 	 	\bottomrule
\end{tabular*}
 \end{table*}
 
\subsection{Coarsened Graph Construction}
A coarsened graph can be constructed with the selected seed nodes. 
We first consider the aggregation of non-selected nodes to seed nodes. To preserve the locality, a seed node aggregates the information of non-selected nodes within its $s$-hop neighborhoods in terms of their proximity. In order to sparsify the connection of the coarsened graphs, we utilize the sparsemax introduced in \citep{martins2016softmax}:
\begin{equation}
	S_{n}(v_{l,i})={\rm Sparsemax}(\bm{\hat{r}}_{l,i}),
\end{equation}
with $\bm{\hat{r}_{l,i}}=[\hat{r}_{l,i,1}, \hat{r}_{l,i,2}, \dots, \hat{r}_{l,i,n_l}]$,
\begin{equation}
\hat{r}_{l,i,j}=\left\{
\begin{array}{ll}
 (R_l)_{i,j}   \qquad  (K_{t})_{i,j} > 0 \ {\rm and} \ j  \notin \{\textbf{idx}[k]\}_{k = 1}^{ n_{l+1}} \\
- \inf \qquad \quad  {\rm others,}	
\end{array}\right.
\end{equation}
where $\{\textbf{idx}[k]\}_{k = 1}^{ n_{l+1}}$ denotes the index set of seed nodes.
Correspondingly, the coarsening matrix $\mathcal{C}_l \in \mathbb{R}^{n_{l+1} \times n_{l}}$ indicates the assignment of vertices in the original graph to vertices in the coarsened graph is obtained by:
\begin{equation}\label{e:21}
(C_l)_{ij}=\left\{
\begin{array}{ll}
S_{n}(v_{l,\textbf{idx}(i)})_j  \qquad  j  \notin \{\textbf{idx}[k]\}^{n_{l+1}}_{k= 1}\\
\delta_{ \textbf{idx}(i),j}  \qquad  \qquad \  j  \in \{\textbf{idx}[k]\}^{n_{l+1}}_{k = 1},
\end{array}\right.
\end{equation}
where the delta function $\delta_{ \textbf{idx}[i],j}$ equals $1$ iff $j=\textbf{idx}[i]$.

After the seed nodes selection and the aggregation of non-selected nodes, we  need to reduce the adjacency matrix of the original graph to another one defined on these seed nodes. In this way, we  obtain the connections between nodes in the coarsened graph accordingly and extract multiscale representations with the following layers. For each pair of nodes in the coarsened graph, we construct their connection by taking into consideration the links between subgraphs of the original graph,  each of which consists of a seed node and its associated non-selected nodes, and compute the corresponding new edge weight by weighted aggregation of these connections. Specifically,   for nodes $v_{l+1,p}, v_{l+1,q} \in \mathcal{V}_{l+1}$ in the coarsened graph, the weight of the edge connecting them is computed as $(A_{l+1})_{pq}=\sum_{k=1}^{n_l}\sum_{m=1}^{n_l}(C_{l})_{pk}(A_{l})_{km}(C_{l})_{qm,}$. If  $(A_{l+1})_{pq}=0$, there is no edge between node $v_{l+1,p}$ and node $ v_{l+1,q}$.  In other words,  the adjacency matrix with the connections in the coarsened graph is  obtained as:
\begin{equation}
	A_{l+1}=C_l A_l  {C_l}^{T}.
\end{equation}

Furthermore, we assign the node signal to a seed node as its original node signal together with the aggregation of its associated nodes through a weighted summarization, i.e.,
\begin{equation}
	X_{l+1}=C_l X_l.
\end{equation}
 
Note that, the proposed pooling operator is generic can be interlaced with graph convolution layers as well as other modules to extract hierarchical multiscale representations of graph data to solve a variety of tasks. It is differentiable with learnable parameters $W_{Ql}$  that can be trained together  with other modules of the graph neural network using diverse gradient-based optimization methods.

\section{Experiments}
We evaluate the proposed graph pooling operator and the state-of-the-arts in graph classification tasks.

\begin{table}[t]
\centering
 \caption{  Dataset statistics and property. }\label{t:1}
 \resizebox{.45\textwidth}{!}{  
\begin{tabular}{l|ccccc}
	\toprule
	Method & D\&D &  PROTEINS & NCI1   &NCI109&MUTAGENICITY\\
	\midrule
		Avg $|\mathcal{V}|$ &284.32&39.06&29.87&29.68&30.32\\
		Avg $|\mathcal{E}|$ &715.66&72.82&32.30&32.13&30.77\\
		\#Classes&2&2&2&2&2\\
		\#Graphs & 1178&1113&4110&4127&4337\\
	\bottomrule
\end{tabular}}
\end{table}

\begin{figure*}[htp]
\centering
\subfigure[SAGPool.]{
\begin{minipage}{0.48\linewidth}
\centering
 \includegraphics[width=3in]{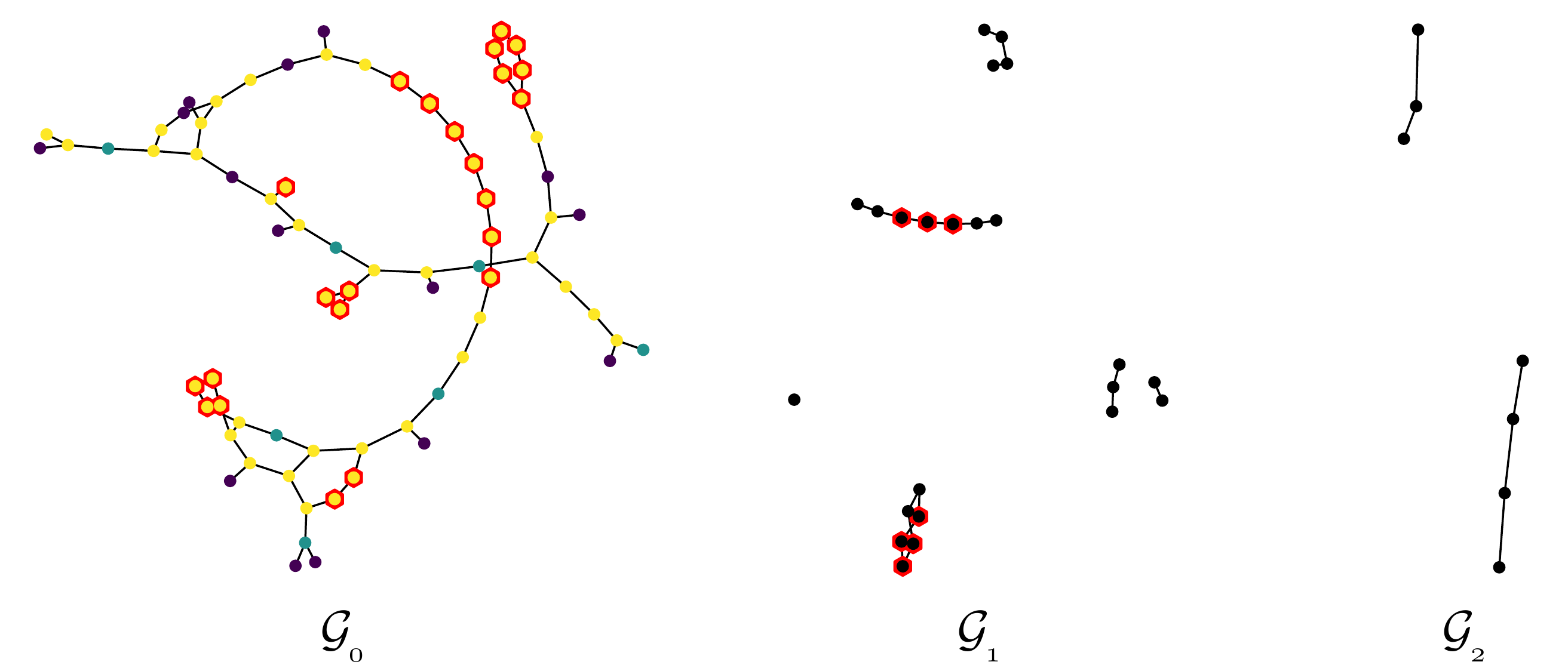}
\end{minipage}
}
\subfigure[gPool.]{
\begin{minipage}{0.48\linewidth}
\centering
 \includegraphics[width=3in]{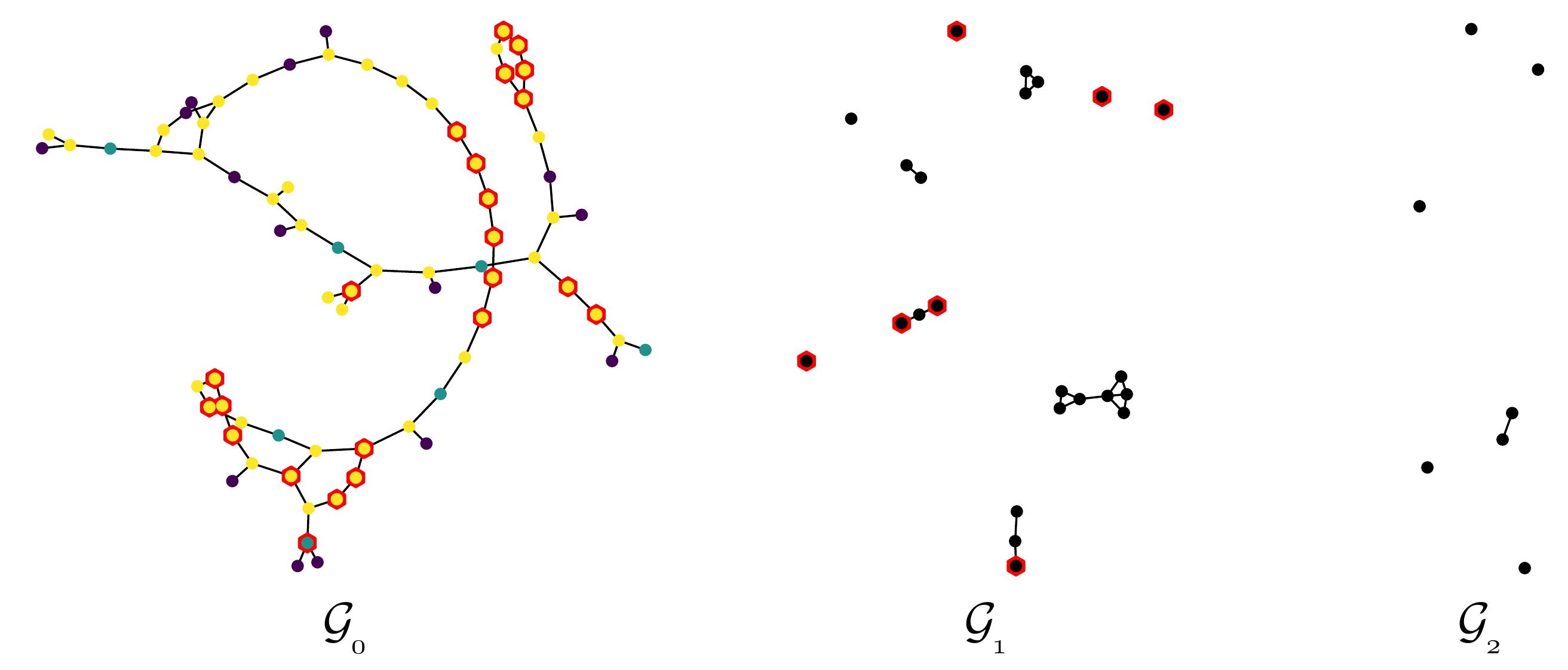}
\end{minipage}
}\\
\subfigure[DIFFPOOL.]{
\begin{minipage}{0.48\linewidth}
\centering
 \includegraphics[width=3in]{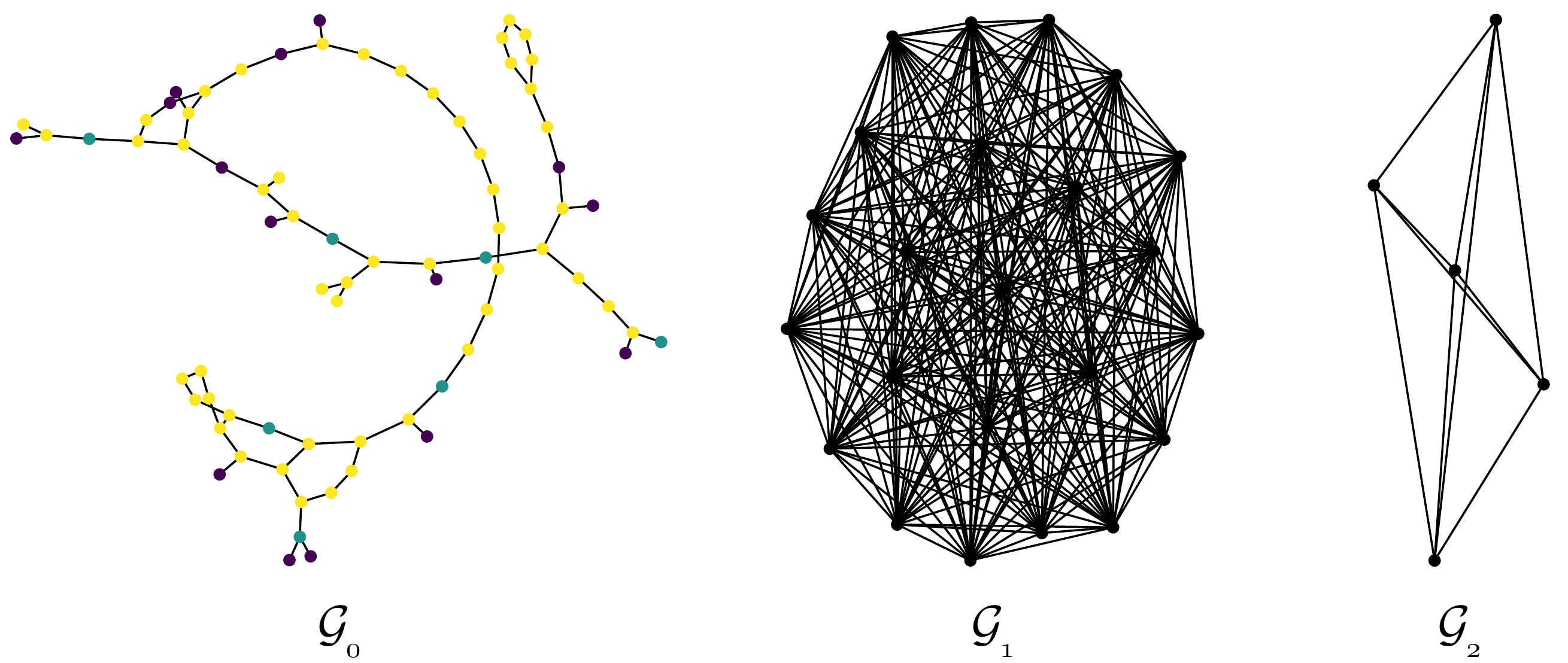}
\end{minipage}
}
\subfigure[ProxPool.]{
\begin{minipage}{0.48\linewidth}
\centering
 \includegraphics[width=3in]{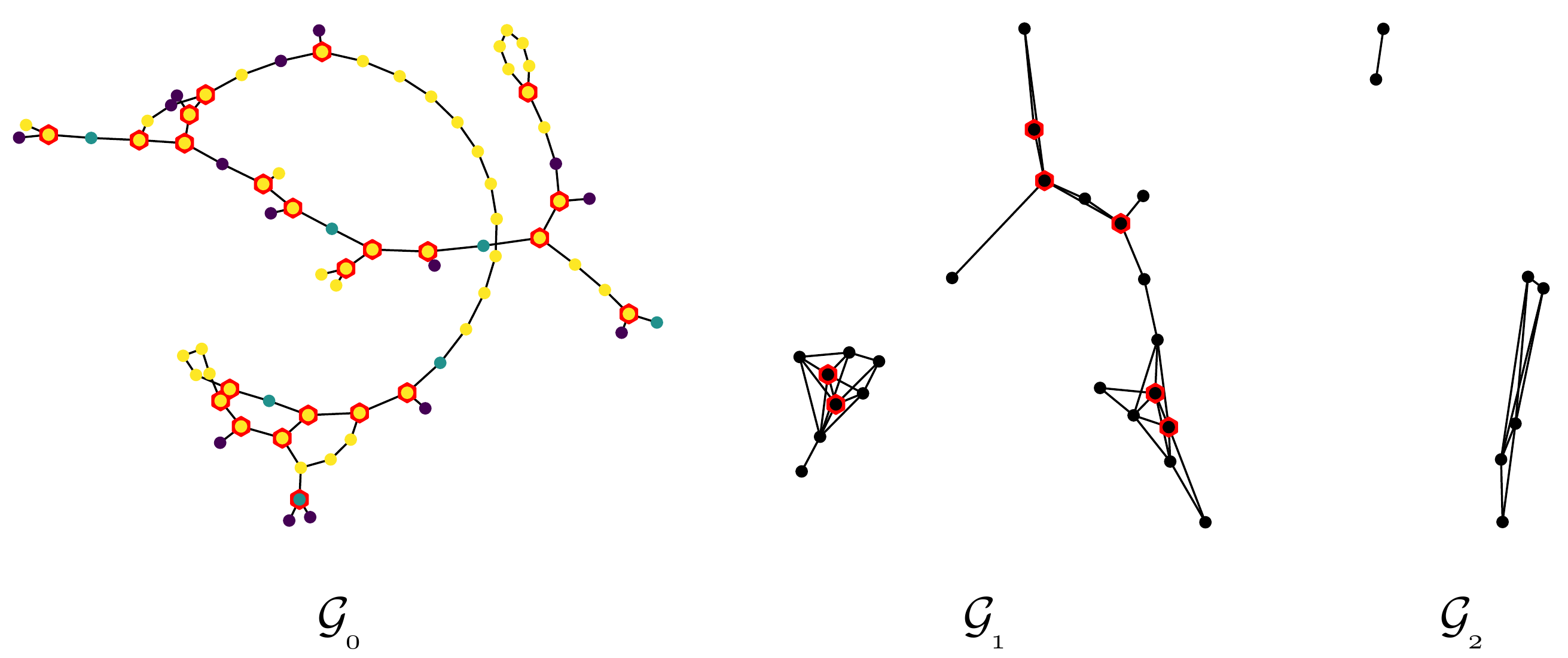}
\end{minipage}
}
\centering
\caption{Comparison of pooling results on a graph of the NCI109 dataset. In each subfigure, we present the original graph and two coarsened graphs generated by the two pooling layers in our architectures. Graph signals with node categories are indicated by different node colors in the original graphs, and seed nodes are denoted by the red hexagons. }\label{fig:2}
\end{figure*}

 \subsection{Experimental Settings} 
\textbf{Datasets.} We follow previous methods  \citep{DBLP:conf/icml/LeeLK19,DBLP:conf/kdd/0001WAT19} to conduct experiments on five large public benchmark graph classification datasets ($|\mathcal{V}|>1000$), including D\&D, PROTEINS, NCI1, NCI109 and MUTAGENICITY\footnote{Datasets could be downloaded from https://ls11-www.cs.tu-dortmund.de/staff/morris/graphkerneldatasets}. Statistics and properties of the datasets are summarized in Table~\ref{t:1}.  Node categorical features are adopted as the node signal.

\textbf{Network architecture.}  We evaluate the proposed pooling operators with the help of deep graph convolution networks. Similar to \citep{ying2018hierarchical,DBLP:conf/icml/LeeLK19,gao2019ipool}, we integrate the proposed pooling operation into the GraphSAGE \citep{hamilton2017inductive} framework. 
In the experiments, the network architecture consists of three convolution layers and two (proposed) pooling layers ([conv-pool]$_{\times2}$-conv), one readout module and one prediction module for all the datasets. We adopt the graph convolution operator with rectified linear unit (ReLU) activation $\sigma$:
\begin{equation}\label{e:1}
	X_{l+1}=\sigma((A_l+I)X_{l}W_{l}),
\end{equation}
where $I$ is the identity matrix and $W_l$ denotes learnable parameters. An $l_{2}$ normalization function is further utilized after each convolution layer to stabilize and accelerate the training process. The pooling operator then follows the convolution layer to coarsen graphs in accordance with the operator proposed in Section~\ref{sec:pool}.  Subsequently, a readout module is adopted to aggregate the graph features at different scales and generate the graph representation $h_{\mathcal{G}}$.
\begin{equation}
    h_{\mathcal{G}}={\rm Concat}(\omega( X_{l})| l=1,3,\dots, K),
\end{equation}
where $\omega(\cdot)$ indicates the node-wise summation and maximum operators to aggregate node information. Finally, a prediction module consisting of two fully connected layers and a softmax layer predicts the class of the graph under study based on the graph representation $h_{\mathcal{G}}$.

\textbf{Configurations.}
According to \citep{DBLP:conf/icml/LeeLK19}, we randomly split each dataset into training, validation and test sets with a ratio of 8:1:1. The trained model achieving best performance on the validation set is selected for test. We conduct 20 random splits for each dataset and report the classification accuracy for test sets. Mean accuracy with standard deviation is used to alleviate the impact of splitting.  

In our experiments, each graph  in a dataset is downsampled with the same pooling ratio $\rho$. The pooling ratio $\rho$ is set as 0.25 on D\&D and 0.3 on others, in consideration of the larger size of the graphs in D\&D. For the network architectures, each convolution layer consists of  64 hidden neurons, and the size of  low-dimension features in the pooling layers obtained through the affine transformation defined in Eq.~(\ref{e:13}) is 16. The proposed models are implemented in Pytorch \citep{paszke2017automatic}, and the models are optimized with the Adam optimizer \citep{DBLP:journals/corr/KingmaB14} with a batch size of 64. The learning rate is 0.001 on all the datasets, except for D\&D using 0.0001. We obtain the following optimal  hyper-parameters through grid search: $\tau \in \{0.1, 1 \}$,  $s \in \{2, 3, 4 \}$, and weight decay $\in \{0, 1e^{-5}, 1e^{-4}\}$.
 
 \textbf{Baselines.} We compare our pooling operator with the recent state-of-the-art graph pooling operators for GCNs. \textbf{DIFFPOOL} \citep{ying2018hierarchical} coarsens graph with an assignment matrix generated by an extra branch of GCNs. Since this branch of GNNs to produce the assignment matrix is predefined and used for all graphs, the sizes of coarsened graphs in a same dataset are the same and are proportional to the maximum number of nodes. We set this proportion as 0.2 so that the average size of coarsened graphs generated by different baselines are similar on most datasets.  \textbf{gPool} \citep{DBLP:conf/icml/GaoJ19} introduces a trainable vector to generate footprint of each node and select nodes accordingly. \textbf{SAGPool}  \citep{DBLP:conf/icml/LeeLK19} first applies attention mechanisms to graph pooling and exploits a graph convolution operation to generate self-attention score. \textbf{EigenPooling} \citep{DBLP:conf/kdd/0001WAT19} relies on spectral clustering algorithms to partition nodes and takes each cluster as a node of the coarsened graph. The node signal is obtained by projecting signals residing on each subgraph to its first $k$ eigenvectors. For fair comparison, all these baselines are re-implemented in the same framework as our pooling operator, except for DIFFPOOL in downsampling as above discussion.

\begin{figure*}[htp]
\centering
\subfigure[The adjacency matrix $A$.]{
\begin{minipage}{0.23\linewidth}
\centering
 \includegraphics[width=1.5in]{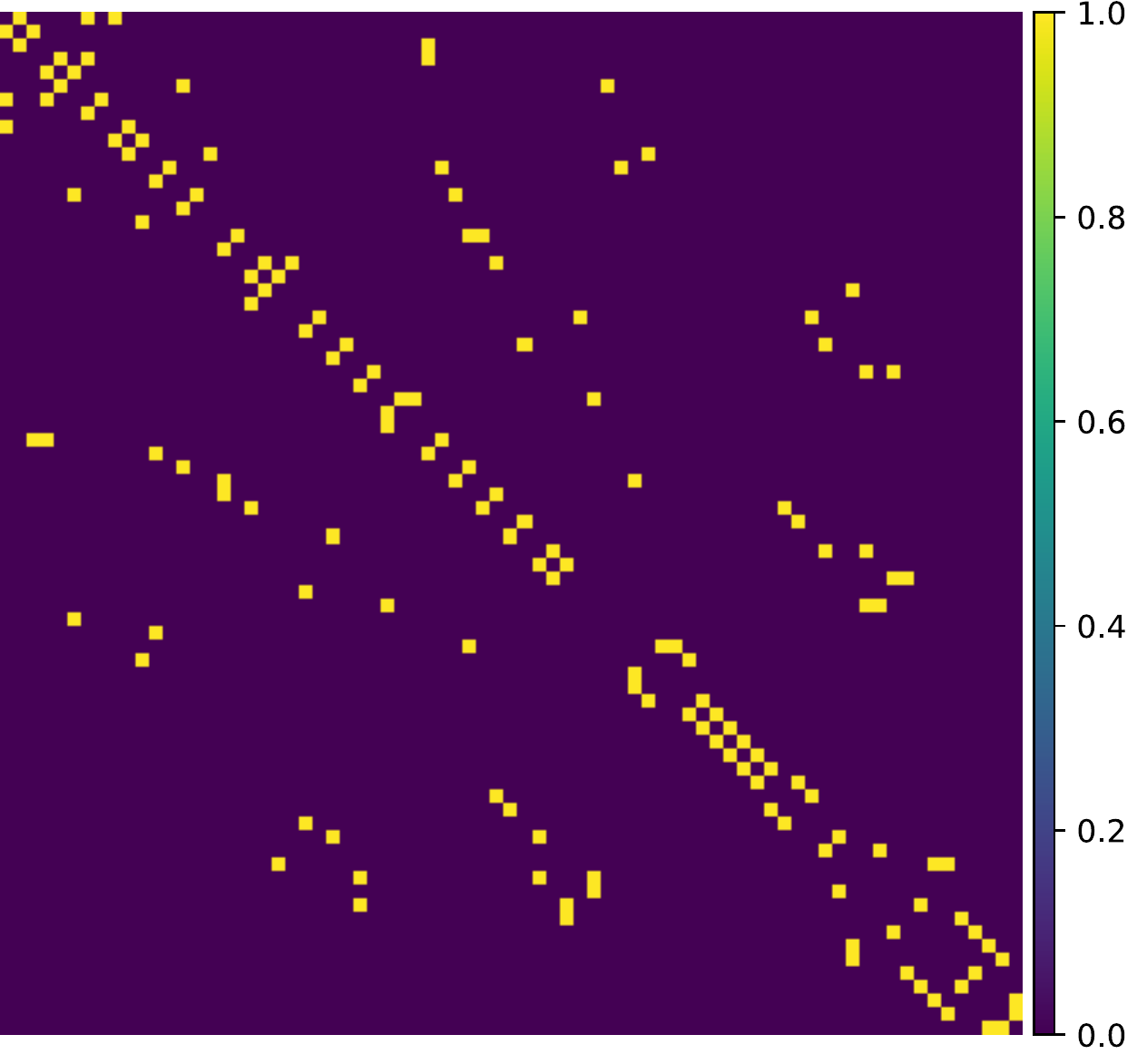}
\end{minipage}
}
\subfigure[The structure kernel $K_t$.]{
\begin{minipage}{0.23\linewidth}
\centering
 \includegraphics[width=1.5in]{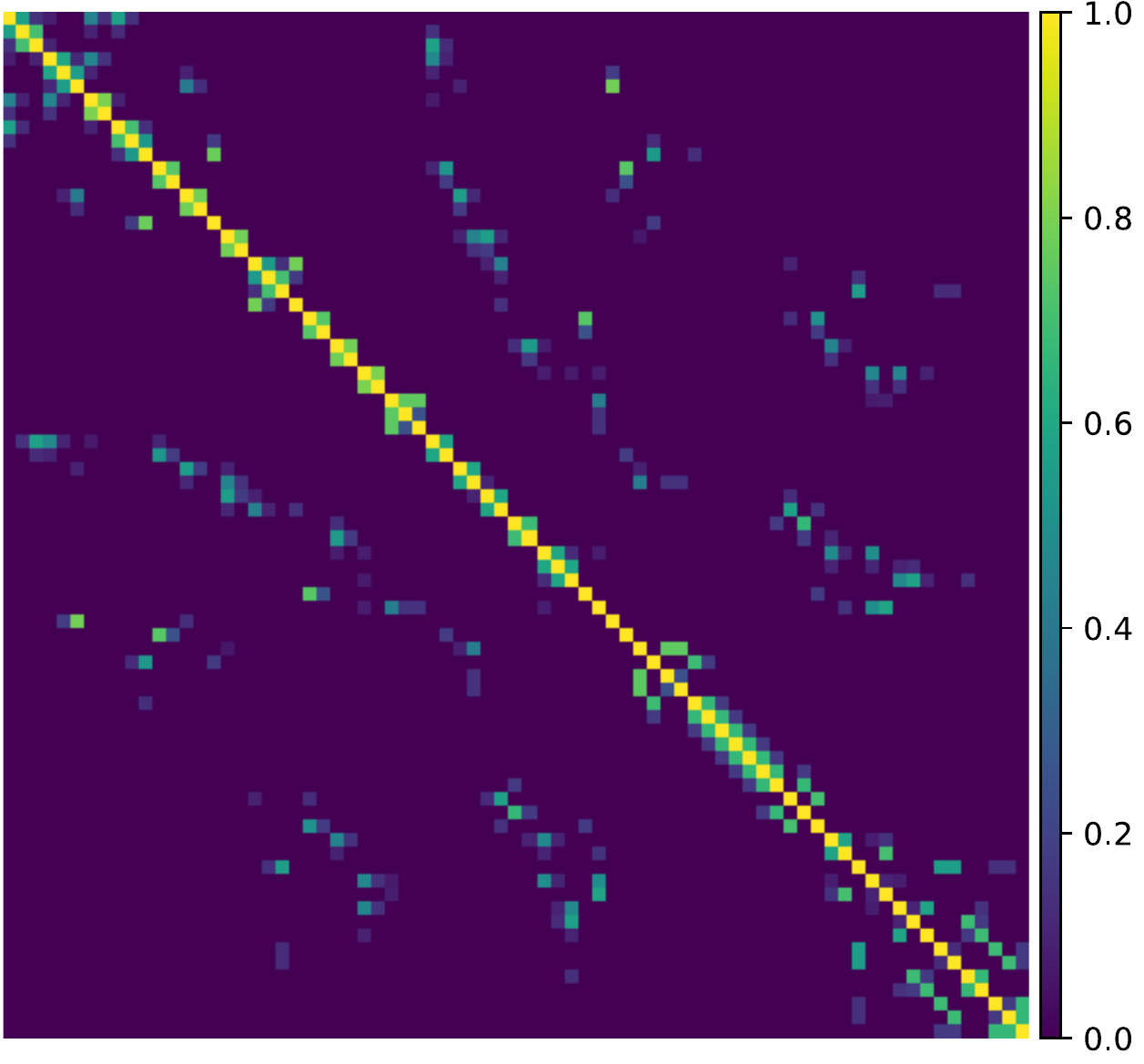}
\end{minipage}
}
\subfigure[The RBF kernel $K_s$.]{
\begin{minipage}{0.23\linewidth}
\centering
 \includegraphics[width=1.5in]{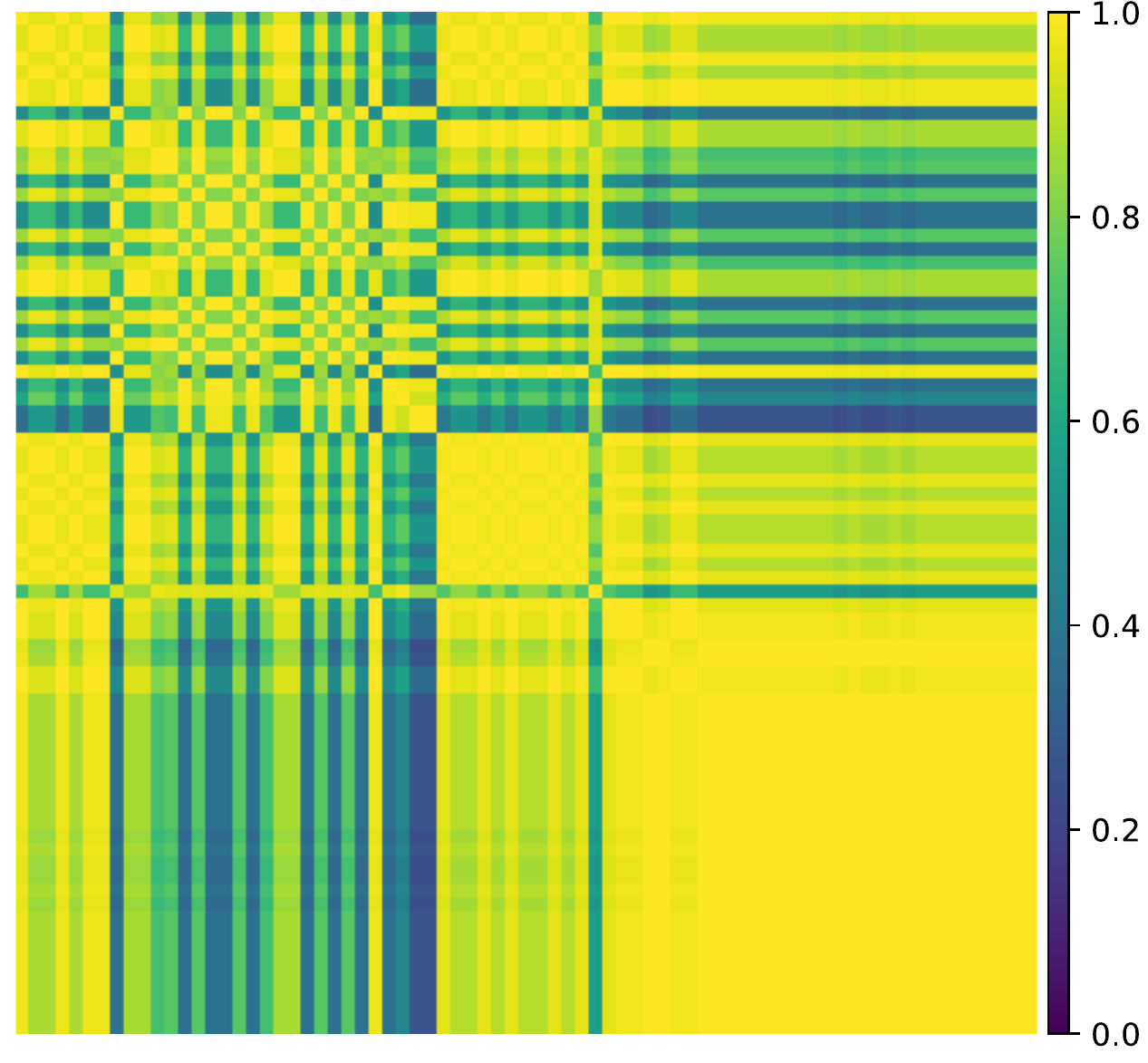}
\end{minipage}
}
\subfigure[Node Proximity $R$.]{
\begin{minipage}{0.23\linewidth}
\centering
 \includegraphics[width=1.5in]{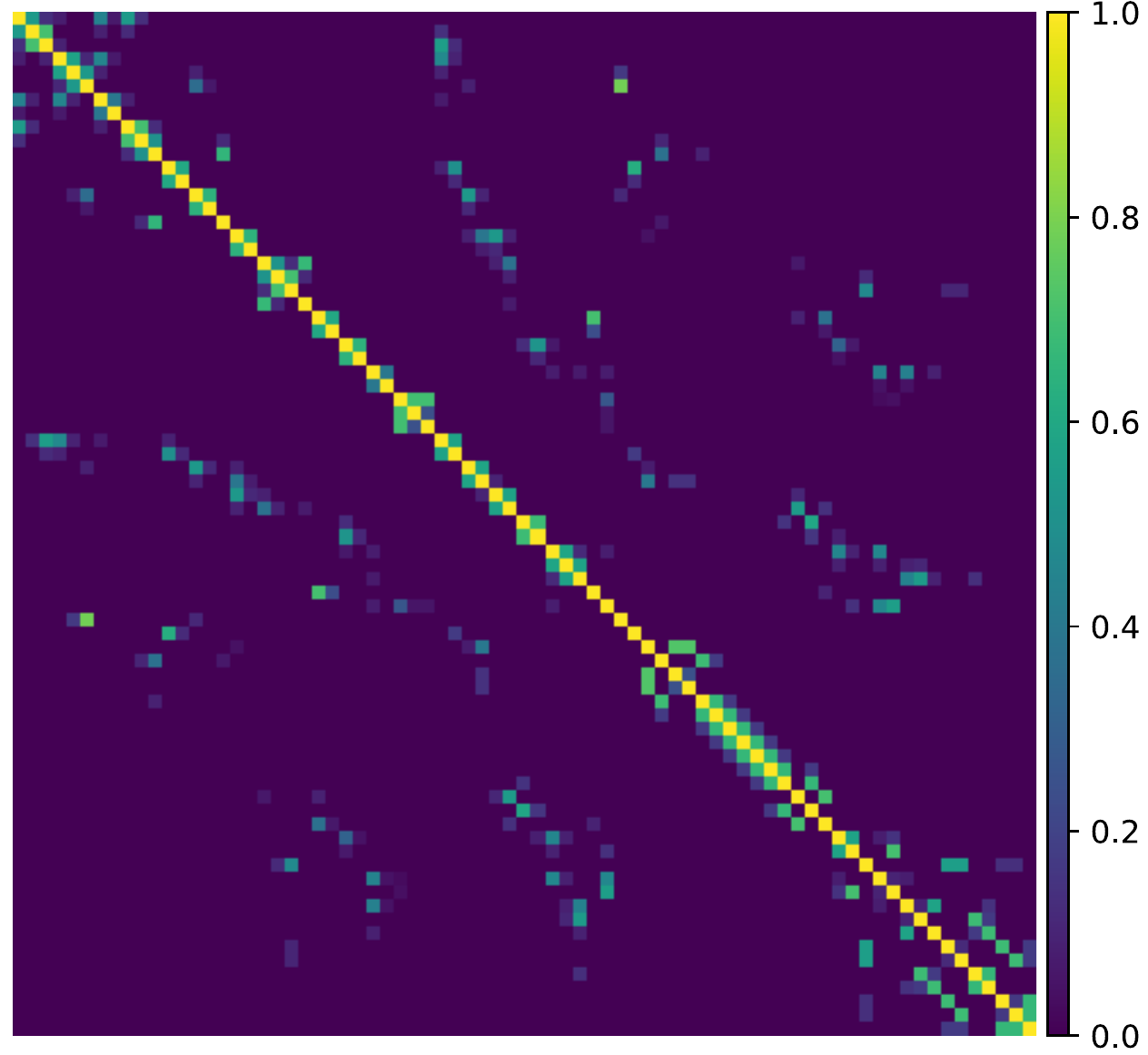}
\end{minipage}
}
\centering
  \caption{Node proximity characteized by different measures.   }\label{fig:4}
\end{figure*}

 \subsection{Ablation Studies}
We evaluate a collection of ablations of the proposed pooling operator for effectiveness of different modules. Besides ProxPool proposed in Section~\ref{sec:prox} and \ref{sec:pool}, we also consider:
\begin{itemize}
  \item \textbf{ProxPool}: The method proposed in Section~\ref{sec:pool}.
   	\item \textbf{ProxPool-NT}: The structure-aware kernel $K_t$ is substituted with the adjacency matrix $A$ to exploit the topology information. The node proximity defined in Eq.~(\ref{e:17}) is reformulated by $R_l=A_l \odot K_s$. Graph downsampling and reduction change accordingly.
  	\item \textbf{ProxPool-NS}: Only the structure-aware kernel $K_t$ is utilized in model node proximity, i.e., $R_l=K_t$. Graph downsampling and reduction change accordingly.
\end{itemize}

\subsection{Results and Discussions}
\begin{figure}[tp]
\centering
\subfigure[DIFFPool.]{
\begin{minipage}{\linewidth}
\centering
 \includegraphics[width=3in]{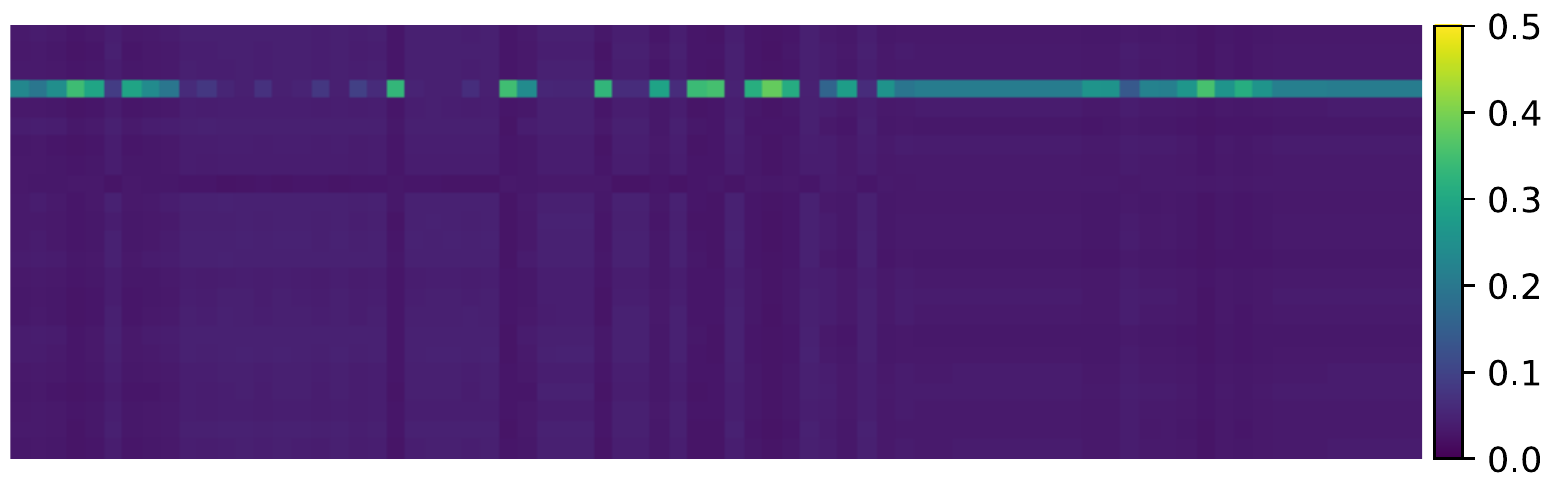}
\end{minipage}
}\\
\subfigure[ProxPool.]{
\begin{minipage}{\linewidth}
\centering
 \includegraphics[width=3in]{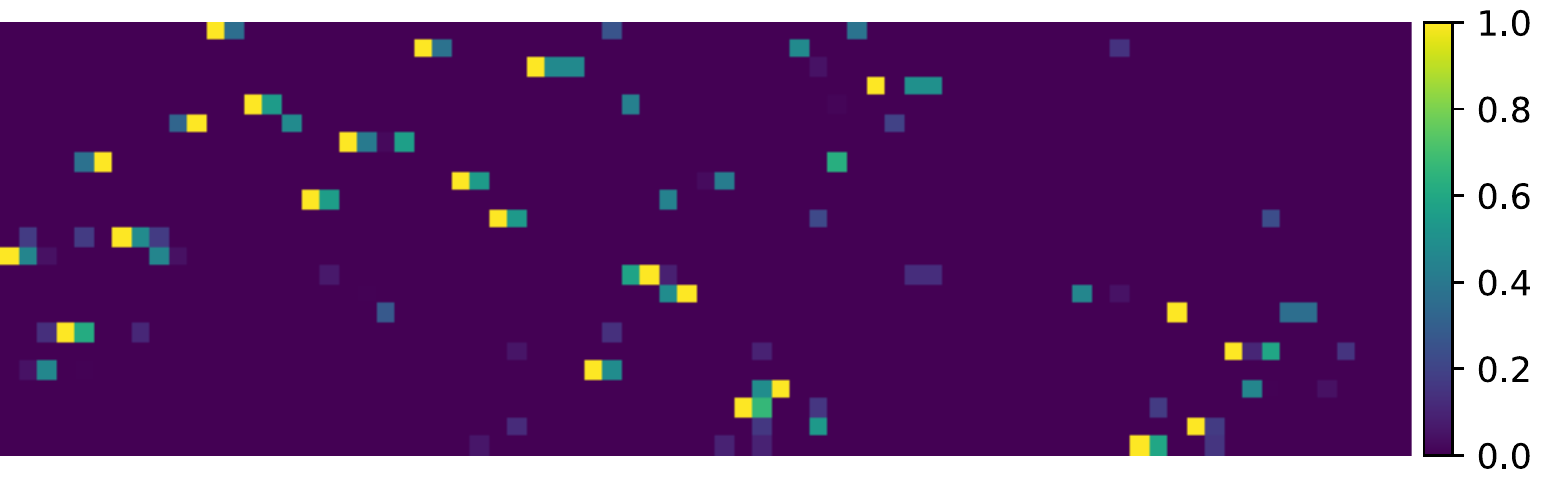}
\end{minipage}
}
\centering
  \caption{Heatmaps of  the assignment matrices of ProxPool and DIFFPOOL on a graph of the NCI109 dataset. Rows are seed nodes, while columns indicate nodes in the original graph.   }\label{fig:3}
  \end{figure}

The comparisons of performance achieved by the baselines, ProxPool and its variants in terms of classification accuracy are presented in Table~\ref{t:2}. The proposed pooling operators outperform all the baseline pooling operators on all the five datasets. Fig.~\ref{fig:2} shows that ProxPool yields better selection of seed nodes than gPool and SAGPool, considering the coupling of seed nodes with non-selected nodes. Furthermore, the coarsened graphs produced by gPool and SAGPool consist of several separate subgraphs with only few nodes. This fact implies that they would impede the information propagation and extraction in the subsequent layers. As shown in Fig.~\ref{fig:2} (c) and Fig.~\ref{fig:3} (a), DIFFPOOL tends to generate complete coarsened graphs with a dense assignment matrix, which leads to the partial loss of the structure information and signal locality. In contrast, Fig.~\ref{fig:3} (b) suggests that ProxPool adopts a sparse assignment matrix to exploit the signal and topology information within $s$-hop neighborhood of seed nodes. Thus, ProxPool is able to balance the connectivity of the structure and the locality of the signal of the coarsened graphs. 

We further study the benefits of the proposed structure-aware kernel. Considering one-hop topology information in modeling the relations between nodes, ProxPool-NT is inferior to ProxPool on all the datasets. As illustrated in Fig.~\ref{fig:4}, the structure-aware kernel captures $s$-hop topology, in addition to the one-hop relationship modeled with the adjacency matrix. These results demonstrate that the proposed structure-aware kernel can sufficiently exploit the topology information of graph data. We also evaluate the node proximity in terms of the graph signal. ProxPool-NS achieves degraded but competitive performance on most datasets, as it only considers the graph topology information. Fig.~\ref{fig:4} shows that the proposed node proximity measure facilitates ProxPool by jointly considering the graph signal and graph topology. 

The computational complexity of ProxPool is dominated by the computation of structure-aware kernel. Thus, its complexity would be $O(|\mathcal{V}_l|^{2.37})$ by optimizing matrix multiplication with the Coppersmith-Winograd algorithm \citep{coppersmith1987matrix}. It can be further reduced with sparse implementation. In contrast, the computational complexity of eigendecomposition of a Laplacian matrix associated with $\mathcal{G}_l$ is $O(|\mathcal{V}_l|^{3})$. These results show that ProxPool leverages the structure-aware kernel to efficiently consider the graph signal and graph topology for pooling. 

\section{Conclusion}
In this paper, we propose a novel graph pooling operator based on the kernel-based measure of node proximity. This measure reconciles the topology and signal information and permits quantitative evaluation of the closeness of arbitrary two nodes within a $s$-hop neighborhood. ProxPool is shown to yield state-of-the-art performance in graph classification. In future, we would employ the proposed node proximity in tasks like node classification and community detection.

\bibliography{egbib.bib}
\bibliographystyle{icml2020}

%\appendix
%\section{Do \emph{not} have an appendix here}
%
%\textbf{\emph{Do not put content after the references.}}
%%
%Put anything that you might normally include after the references in a separate
%supplementary file.
%
%We recommend that you build supplementary material in a separate document.
%If you must create one PDF and cut it up, please be careful to use a tool that
%doesn't alter the margins, and that doesn't aggressively rewrite the PDF file.
%pdftk usually works fine. 
%
%\textbf{Please do not use Apple's preview to cut off supplementary material.} In
%previous years it has altered margins, and created headaches at the camera-ready
%stage. 
%%%%%%%%%%%%%%%%%%%%%%%%%%%%%%%%%%%%%%%%%%%%%%%%%%%%%%%%%%%%%%%%%%%%%%%%%%%%%%%
%%%%%%%%%%%%%%%%%%%%%%%%%%%%%%%%%%%%%%%%%%%%%%%%%%%%%%%%%%%%%%%%%%%%%%%%%%%%%%%

\end{document}